\pdfoutput=1
\documentclass[11pt,a4paper]{article}
\usepackage[hyperref]{acl2020}
\usepackage{times}
\usepackage{latexsym}

\usepackage{graphicx}
\usepackage{url}
\usepackage{subcaption}
\usepackage{amsmath,bm}
\usepackage{paralist}
\usepackage{hyperref}
\usepackage{multicol}
\usepackage{color}
\usepackage{booktabs}
\usepackage{xspace}
\usepackage{textcomp}
\usepackage{amssymb}
\usepackage{enumitem}
\setitemize{noitemsep,topsep=0pt,parsep=0pt,partopsep=0pt}
\usepackage[activate={true,nocompatibility}, kerning=true,spacing=true, stretch=10,shrink=30]{microtype}
\microtypecontext{spacing=nonfrench}

\def\vec#1{\ensuremath{\boldsymbol{{#1}}}}

\aclfinalcopy %

\newcommand{\interalia}[1]{\citep[\emph{inter alia}]{#1}}

\newcommand{\coherent}{\textsc{coherent}\xspace}
\newcommand{\diffsum}{\textsc{diffsum}\xspace}
\newcommand{\segendpoints}{\textsc{endpoint}\xspace}
\newcommand{\attn}{\textsc{attn}\xspace}
\newcommand{\segmax}{\textsc{max}\xspace}
\newcommand{\segavg}{\textsc{avg}\xspace}

\newcommand{\bert}{BERT\xspace}

\newcommand{\roberta}{RoBERTa\xspace}
\newcommand{\robertalarge}{RoBERTa-large\xspace}

\newcommand{\nlw}{\hspace{0.15in}-mix\_wt}

\title{A Cross-Task Analysis of Text Span Representations}%

\author{Shubham Toshniwal,
Haoyue Shi,
Bowen Shi,
Lingyu Gao,
Karen Livescu,
Kevin Gimpel\\
Toyota Technological Institute at Chicago\\[0.5em]
\small{\texttt{\{shtoshni, freda, bshi, lygao, klivescu, kgimpel\}@ttic.edu }}\\
}

\date{}

\begin{document}
\maketitle
\begin{abstract}
Many natural language processing (NLP) tasks involve reasoning with textual spans, including question answering, entity recognition, and coreference resolution. While extensive research has focused on functional architectures for representing words and sentences, there is less work on representing arbitrary spans of text within sentences. In this paper, we conduct a comprehensive empirical evaluation of six span representation methods using eight pretrained language representation models across six tasks, including two tasks that we introduce. We find that, although some simple span representations are fairly reliable across tasks, in general the optimal span representation varies by task, and can also vary within different facets of individual tasks. We also find that the choice of span representation has a bigger impact with a fixed pretrained encoder than with a fine-tuned encoder.

\end{abstract}

\section{Introduction}

Fixed-dimensional span representations are often used as a component in recent models for a number of
natural language processing (NLP) tasks, such as question answering \citep{lee-etal-2016-learning,seo-etal-2019-real}, coreference resolution \citep{lee-etal-2017-end} and constituency parsing \interalia{stern-etal-2017-minimal,kitaev-klein-2018-constituency,kitaev-etal-2019-multilingual}.
Such models initialized
with contextualized word or token embeddings \citep{peters2018deep,devlin2019bert} have achieved new state-of-the-art results for these tasks \citep{kitaev-etal-2019-multilingual,joshi-etal-2019-bert}.

\begin{figure}[t]
    \begin{subfigure}{0.2\textwidth}
        \includegraphics[height=0.16\textheight]{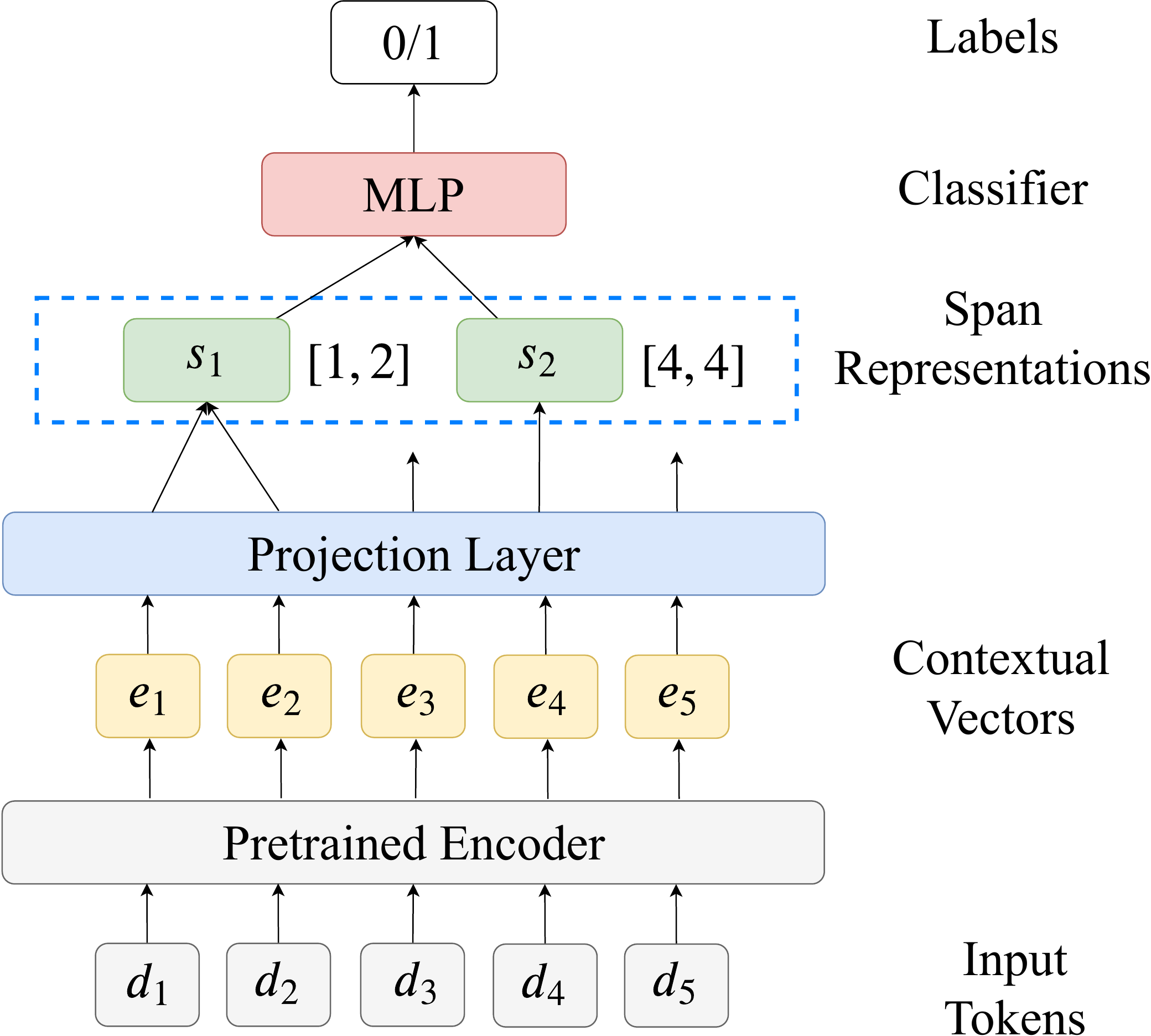}
        \caption{Probing model for two-span tasks. This model can be used to decide whether two spans (here $[1,2]$ and $[4,4]$) are coreferent or not.}
    \end{subfigure}%
    \hspace{0.5in}%
    \begin{subfigure}{0.2\textwidth}
        \includegraphics[height=0.16\textheight]{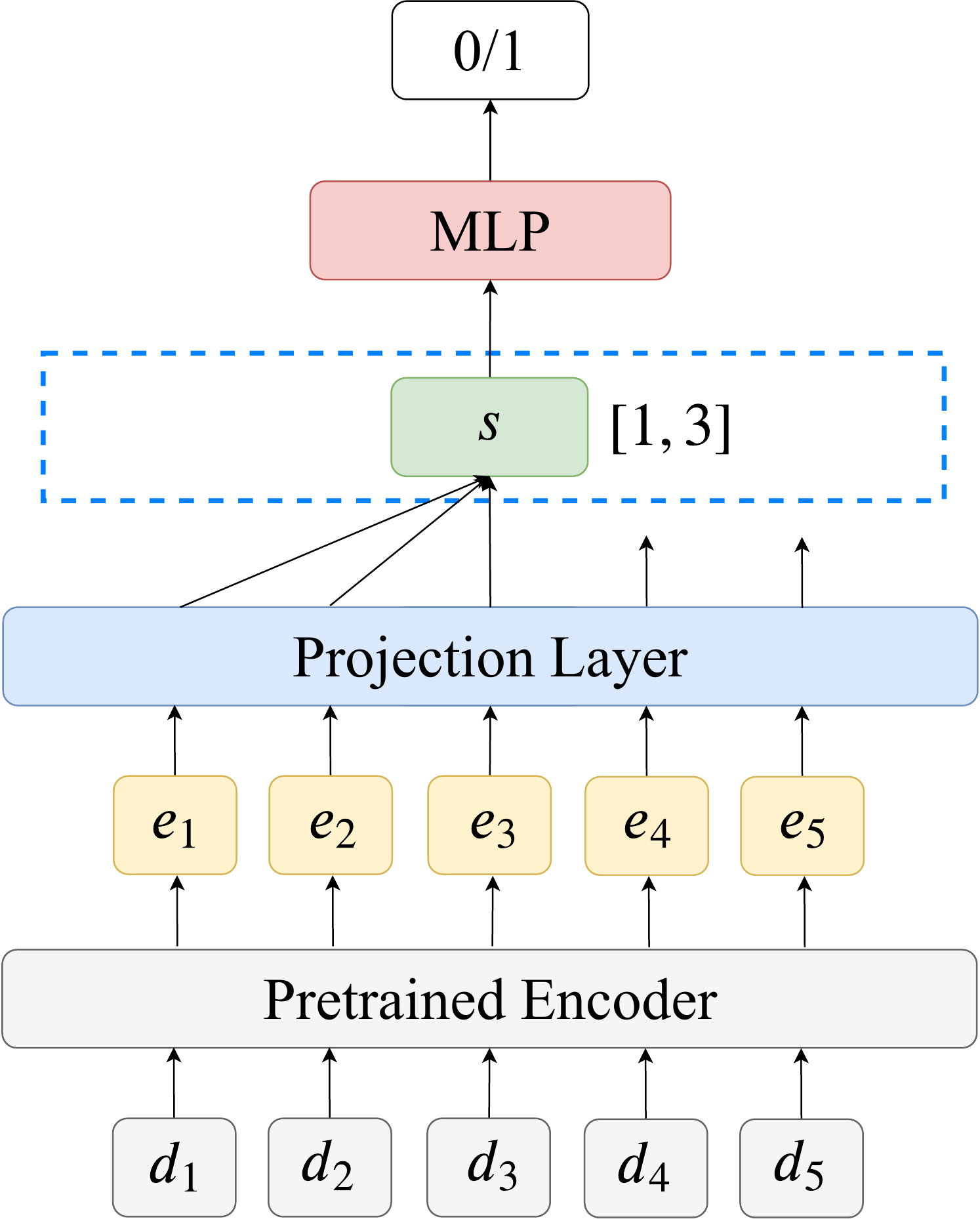}
        \caption{Probing model for single-span tasks. This model can be used to decide whether a span (here $[1,3]$) refers to a constituent or not.}
    \end{subfigure}%
\caption{
The models are very similar to that of \citet{tenney2019} but we explicitly separate the span representation part into a projection step followed by a choice among several span representation methods.}
\label{fig:model}
\end{figure}

Span representations are usually computed from token embeddings, such as word embeddings \cite{pennington-etal-2014-glove} or byte pair encoding \cite{gage1994new} based embeddings \cite{sennrich-etal-2016-neural}.
Since spans can have arbitrary length (i.e., number of tokens), fixed-dimensional span representations involve some form of (parameterized) pooling of the token representations.
Existing models typically pick a
\textit{span representation} method (dashed boxes in Figure~\ref{fig:model}) that works well for the task(s) of interest.  However, a comprehensive evaluation comparing various %
{span representation} methods across tasks is still lacking.\footnote{Code available at \url{https://github.com/shtoshni92/span-rep}}

In this work, we systematically compare and analyze a wide range of
{span representations} (Section~\ref{sec:pooling_models})
by probing the representations via various NLP tasks, including constituent detection, constituent labeling, named entity labeling, semantic role labeling, mention detection, and coreference arc prediction (Section~\ref{sec:task}).
All of the tasks we consider naturally involve span representations. Similar comparisons are done %
by ~\citet{tenney2019}, where they use this probing approach to compare several pretrained contextual embedding models, while keeping the span representation method fixed to self-attentive pooling \citep{lin2017structured,lee-etal-2017-end}.
Here we vary both the choice of contextualized embedding models (among \bert~\citep{devlin2019bert}, \roberta~\citep{liu2019roberta}, XLNet~\citep{yang2019xlnet}, and SpanBERT~\citep{joshi2019spanbert}) and the span representation methods.
By analyzing the performance of each span representation method for multiple tasks, we aim to uncover the importance of choice of span representation.

We follow the ``edge probing" setup of \citet{tenney2019} and introduce two new tasks to this setup, namely constituent detection and mention detection, which complement the constituent labeling and coreference arc prediction tasks, respectively, that are part of the original setup.
For the full-scale comparison, we follow the original setup and keep the pretrained token representation models fixed and learning only layer weights and additional task-specific parameters on top of the weighted pretrained representations.
We also conduct a small-scale study to compare the effect of fine-tuning on different span representations in terms of their relative ordering and for comparison with their non fine-tuned counterparts.

Overall, we find that the behavior of span representations tends to pattern according to whether they are based on information at the span boundaries versus using the entire span content.
With frozen pre-trained models, we find that the choice of span representation is more important than the choice of base token embedding model.
For the fine-tuned case, choice of span representation still has an impact on performance though it's much less pronounced in comparison to the non fine-tuned case.
In general, a span representation that takes the max over time is a reliable choice across tasks, although the best-performing model can vary greatly between tasks. %

\section{Task Formulation}
\label{sec:task}
We borrow four probing tasks applied by \citet{tenney2019}, namely constituent labeling, named entity labeling, semantic role labeling, and coreference arc prediction.
We also introduce two new tasks: constituent detection and mention detection.
The specific tasks are described below. \\[0.2cm]
\textbf{Constituent labeling} is the task of predicting the non-terminal label (e.g., noun phrase or verb phrase) for a span corresponding to a constituent. \\[0.2cm]
\textbf{Constituent detection} is the task of determining whether a span of words corresponds to a constituent
(i.e., a nonterminal node) in the constituency parse tree of the input sentence.
We introduce this task as a complementary to the task of constituent labeling, to further evaluate the syntax ability of the span representation methods. \\[0.2cm]
\textbf{Named entity labeling (NEL)} is the task of predicting the entity type of a given span corresponding to an entity. For example, does the span ``German" in some context refer to people, organization, or language.\\[0.2cm]
\textbf{Semantic role labeling (SRL)} is concerned with predicting the semantic roles of phrases in a sentence. In this probing task the locations of the predicate and its argument are given, and the goal is to classify the argument into its specific semantic roles (\texttt{agent}, \texttt{patient}, etc.).  \\[0.2cm]
\textbf{Mention detection} is the task of predicting whether a span represents a mention of an entity or not. For example, in the sentence \textit{``Mary goes to the market"}, the spans  \textit{``Mary"} and  \textit{``the market"} refer to mentions while all other spans are not mentions. The task is similar to named entity recognition \citep{tjong-kim-sang-de-meulder-2003-introduction}, but the mentions are not limited to named entities. We introduce this task as it is the first step for coreference resolution \citep{pradhan-etal-2012-conll}, if the candidate mentions are not explicitly given. %

\textbf{Coreference arc prediction} is the task of predicting whether a pair of spans refer to the same entity or not. For example, in ``John is his own enemy", ``John" and ``his" refer to the same person.

\section{Model}
\label{sec:model}
In this section, we first briefly describe the probing model, which is borrowed from \citet{tenney2019} with the extension to different span representations (Figure~\ref{fig:model}), followed by details of the various span representation methods we compare in this work.

\subsection{Probing Model}
The input to the model is a sentence $\mathbf{d} = \{d_1, \cdots, d_{T}\}$ where the $d_i$ are tokens (produced by a tokenizer specific to a given choice of encoder).  The sentence is first passed through a fixed, pretrained encoder, such as BERT, followed by a learned projection layer to obtain contextualized token embeddings $\{\vec{e}_1, \cdots, \vec{e}_T\}$. These embeddings are then fed to span representation modules to get fixed-dimensional contextual span embeddings. Finally, the span embeddings are fed into a two-layer MLP followed by a sigmoid layer to predict the labels.
For multiclass probing tasks with $|\mathcal{L}|$ labels, the predictions are made independently with separate MLPs per label resulting in a $[0, 1]^{|\mathcal{L}|}$ vector.
 Finally, some tasks involve a single span, whereas others (coreference, semantic role labeling) involve two spans; in the latter case, the MLP takes as input the concatenation of the representations corresponding to the two spans.

\subsection{Span Representation Methods}
\label{sec:pooling_models}
Given a span $s = [i, j]$ and its corresponding contextualized word embeddings $[\vec{e}_i, \cdots, \vec{e}_j]$, where $\vec{e}_k \in \mathbb{R}^d$, a span representation module outputs a fixed-dimensional span representation $\vec{s}_{ij}$.
Below we describe the various span representation methods compared in this work, including the attention pooling used by \citet{tenney2019}. \\[0.2cm]
\textbf{Average pooling} is a simple average of the contextualized word embeddings in the span window:
$$\vec{s}_{ij} = \frac{1}{j - i + 1}\sum_{k=i}^j \vec{e}_k$$ \\[0.2cm]
\textbf{Attention pooling} or self-attention pooling is a learned weighted average over the contextualized token embeddings in the span:
\begin{align*}
    \alpha_k &= \vec{v} \cdot \vec{e}_k; ~~~a_k = \text{softmax}(\vec{\alpha})_k\\
    \vec{s}_{ij} &= \sum_{k=i}^j a_k \cdot \vec{e}_k
\end{align*}
where $\vec{v}$ is a learned parameter vector. This pooling method is a popular choice for many NLP tasks \citep{lee-etal-2017-end,lin2017structured},  and is the one used by \citet{tenney2019}. \\[0.2cm]
\textbf{Max pooling}\label{sec:max_pool} takes the maximum value over time for each
dimension of the input contextualized embedding vectors within the span.
Max pooling has been frequently used to obtain fixed-dimensional sentence representations for classification tasks~\cite{Collobert:2011:NLP:1953048.2078186, hashimoto-etal-2017-joint, conneau-etal-2017-supervised}.  \\[0.2cm]
\textbf{Endpoint} is a simple concatenation of the endpoints of the span: $\vec{s}_{ij} = [\vec{e}_i; \vec{e}_j]$. This is a popular choice for representing answer spans~\cite{lee-etal-2016-learning} in extractive question-answering tasks such as SQuAD~\cite{rajpurkar-etal-2016-squad}. Note that in this case $\vec{s}_{ij} \in \mathbb{R}^{2d}$.  \\[0.2cm]
\textbf{Diff-Sum} is a
variant of endpoint where we concatenate the sum and difference of the span endpoints: $\vec{s}_{ij} = [\vec{e}_j + \vec{e}_i; \vec{e}_j - \vec{e}_i]$. Diff-sum and its
close variants have been used in parsing and SRL
~\cite{stern-etal-2017-minimal, ouchi-etal-2018-span}. As in endpoint, $\vec{s}_{ij} \in \mathbb{R}^{2d}$.   \\[0.2cm]
\textbf{Coherent} is a span representation proposed by~\citet{seo-etal-2019-real} for indexing phrases in a query-agnostic manner for question answering.
First, the endpoints of the span are split into four parts: $$\vec{e}_i = [\vec{e}_i^1; \vec{e}_i^2; \vec{e}_i^3; \vec{e}_i^4]$$
where $\vec{e}_i^1, \vec{e}_i^2 \in \mathbb{R}^a$ and $\vec{e}_i^3, \vec{e}_i^4 \in \mathbb{R}^b$, and therefore $2a + 2b = d$. The endpoints are then combined as:
$$\vec{s}_{ij} = [\vec{e}_i^1; \vec{e}_j^2; \vec{e}_i^3\cdot\vec{e}_j^4]$$
where the last term $\vec{e}_i^3\cdot\vec{e}_j^4$ is referred to as the {\it coherence} term; hence the name ``coherent" (assigned by us). In this case, $\vec{s}_{ij} \in \mathbb{R}^{2a + 1}$ where $2a + 1 < d$ since $2a + 2b = d$.\footnote{\citet{seo-etal-2019-real} used $a$=480 and $b$=32 for $1024$-dimensional BERT-large embeddings. We use the same proportions for the projected contextual embeddings.}

\section{Experimental Setup}

\begin{table*}[htbp]
\setlength{\tabcolsep}{4pt}
\centering
\begin{tabular}{l l l l l}
\toprule
Task & Task Type & $|\mathcal{L}|$  %
 & Total Examples\\
\midrule
Constituent labeling & Syntactic &  30 %
 &  1.9M / 255K / 191K \\
Constituent detection & Syntactic &  2 %
 & \\
NEL & Semantic                      & 18  %
 & 128K / 20K / 13K\\
SRL & Semantic/Syntactic &  66 %
 & 599K / 83K / 62K \\
Mention detection & Syntactic &  2 %
& 387K/ 49K/ 48K\\
Coreference arc prediction & Semantic & 2 %
 & 208K / 27K / 28K \\
\bottomrule
\end{tabular}
\caption{
Dataset statistics of all the six tasks.
}
\label{tab:data_stats}

\end{table*}

\subsection{Implementation details}
All input strings are passed through contextual encoder models to obtain an embedding for each token.
The weighted average of outputs from all layers are used as the token representation \citep{tenney2019}.
We investigate four pre-trained models: \bert \citep{devlin2019bert}, \roberta \citep{liu2019roberta}, SpanBERT \cite{joshi2019spanbert}, and XLNet \cite{yang2019xlnet}. Each has both ``base'' and ``large'' variants,
and we experiment with both.
Since some of the models, such as XLNet, only have cased versions, we use the cased version for all models.
We use the HuggingFace \citep{Wolf2019HuggingFacesTS} implementation of the four models, which is based on PyTorch \cite{pytorch}.

Embeddings are first projected down to 256 dimensions.\footnote{For SRL we use separate projection matrices for the two spans involved in the task, as the two spans may require different types of information to be extracted.  For all of the other tasks, a single projection matrix is used.} For each span, a pooling operation (one of the methods from Section~\ref{sec:pooling_models}) is then applied to its sequence of projected vectors to obtain a fixed-length representation for the span. The span representations are concatenated (if there are more than one) and fed into a two-layer MLP followed by a sigmoid output layer. The two-layer MLP is a stack of a linear layer, a non-linear layer with $\tanh$ activations, layer normalization, dropout (0.3 zeroing probability), and a second linear layer. The hidden dimension of the MLP is 256. Models are trained by minimizing binary cross-entropy loss against the set of true labels. Though some tasks (e.g., SRL) are multi-class classification, we make predictions for each label independently, which facilitates analysis on individual labels or label groups.%

\begin{figure*}[t!] %
\begin{subfigure}{0.42\textwidth}
 \includegraphics[height=0.18\linewidth, height=2.6in]{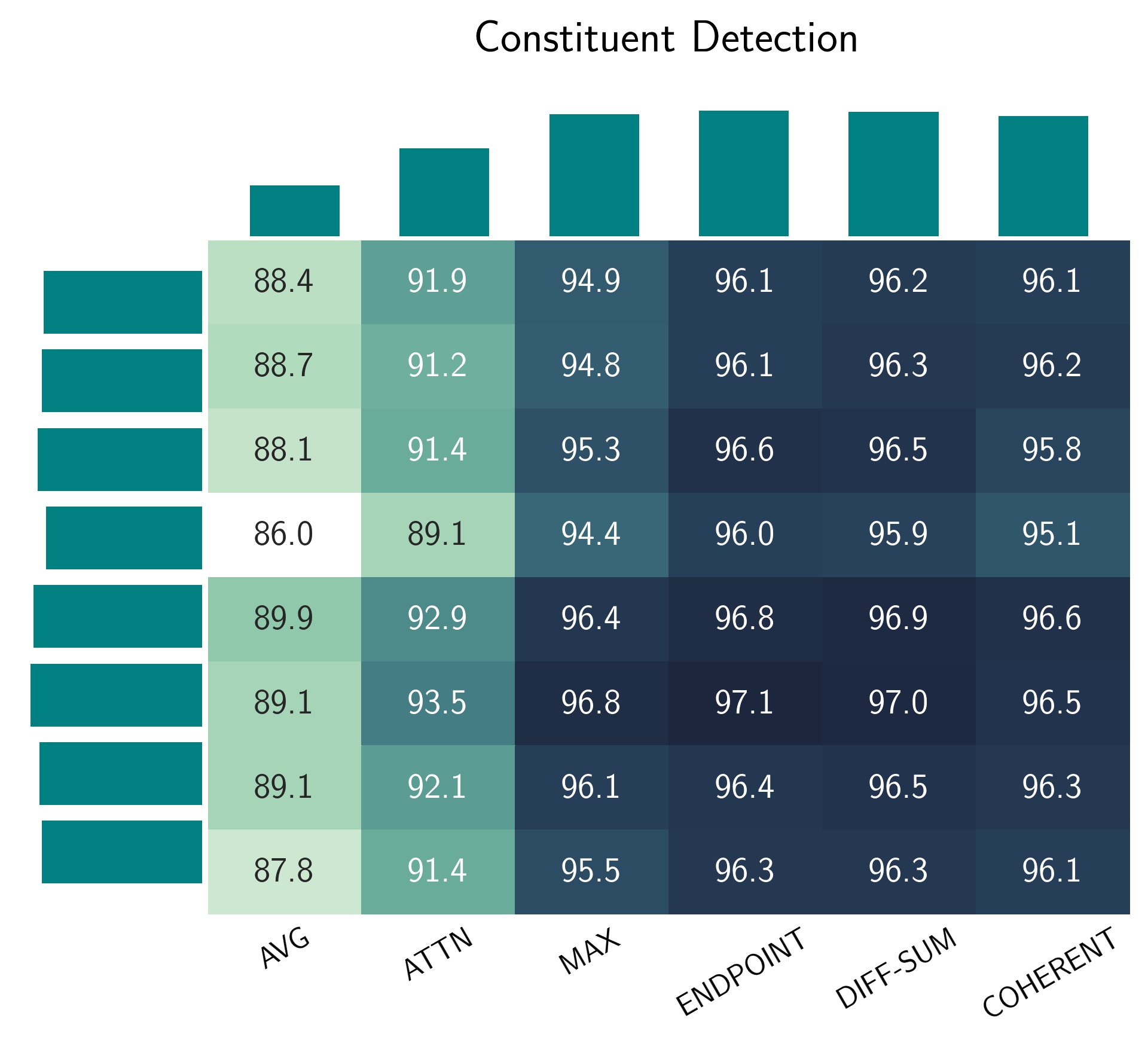}
\end{subfigure}\hspace*{\fill}
\begin{subfigure}{0.56\textwidth}
\includegraphics[height=0.18\linewidth,  height=2.6in]{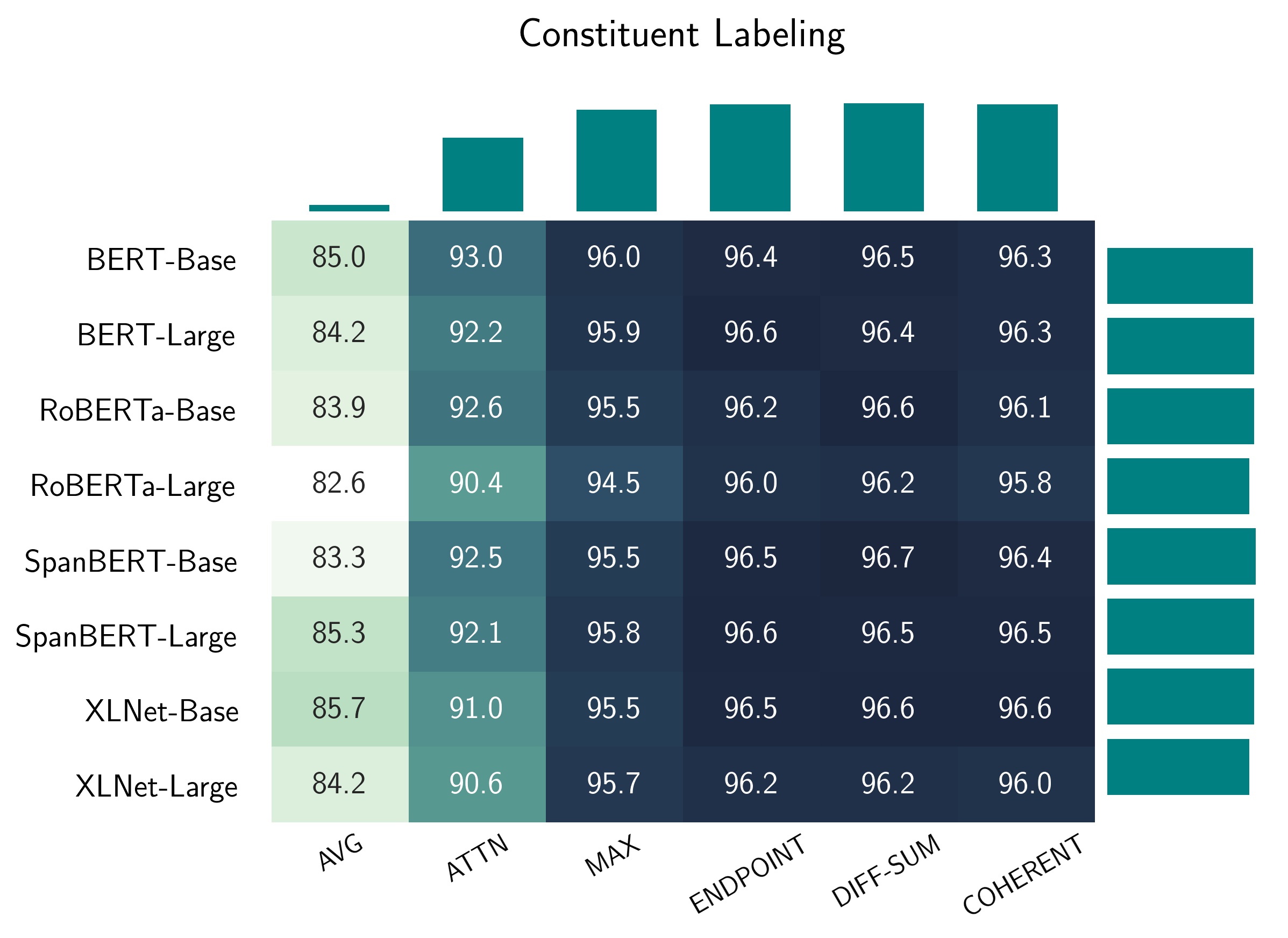}
\end{subfigure}

\medskip
\begin{subfigure}{0.42\textwidth}
\includegraphics[height=0.18\linewidth, height=2.6in]{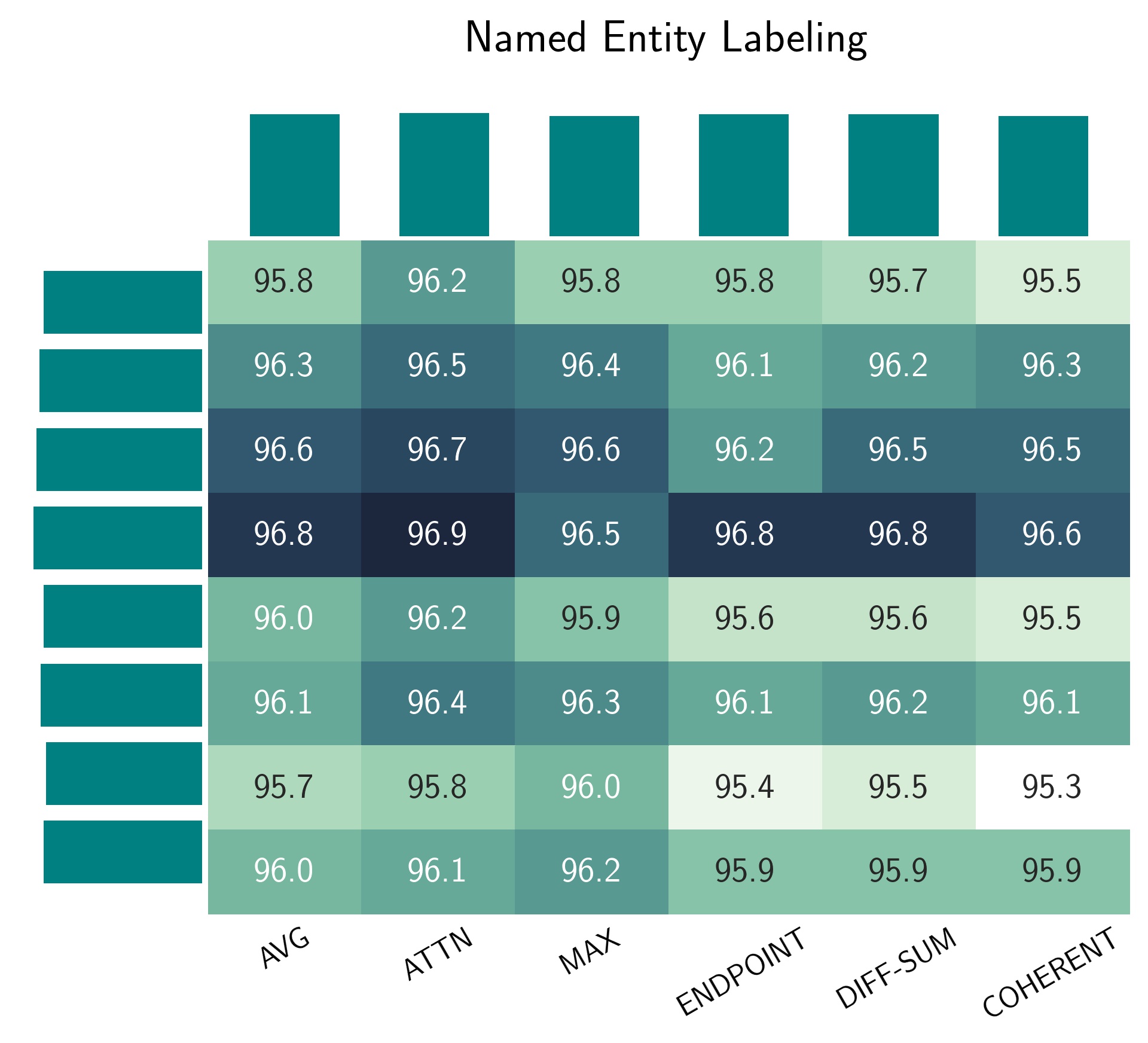}
\end{subfigure}\hspace*{\fill}
\begin{subfigure}{0.56\textwidth}
\includegraphics[height=0.18\linewidth,  height=2.6in]{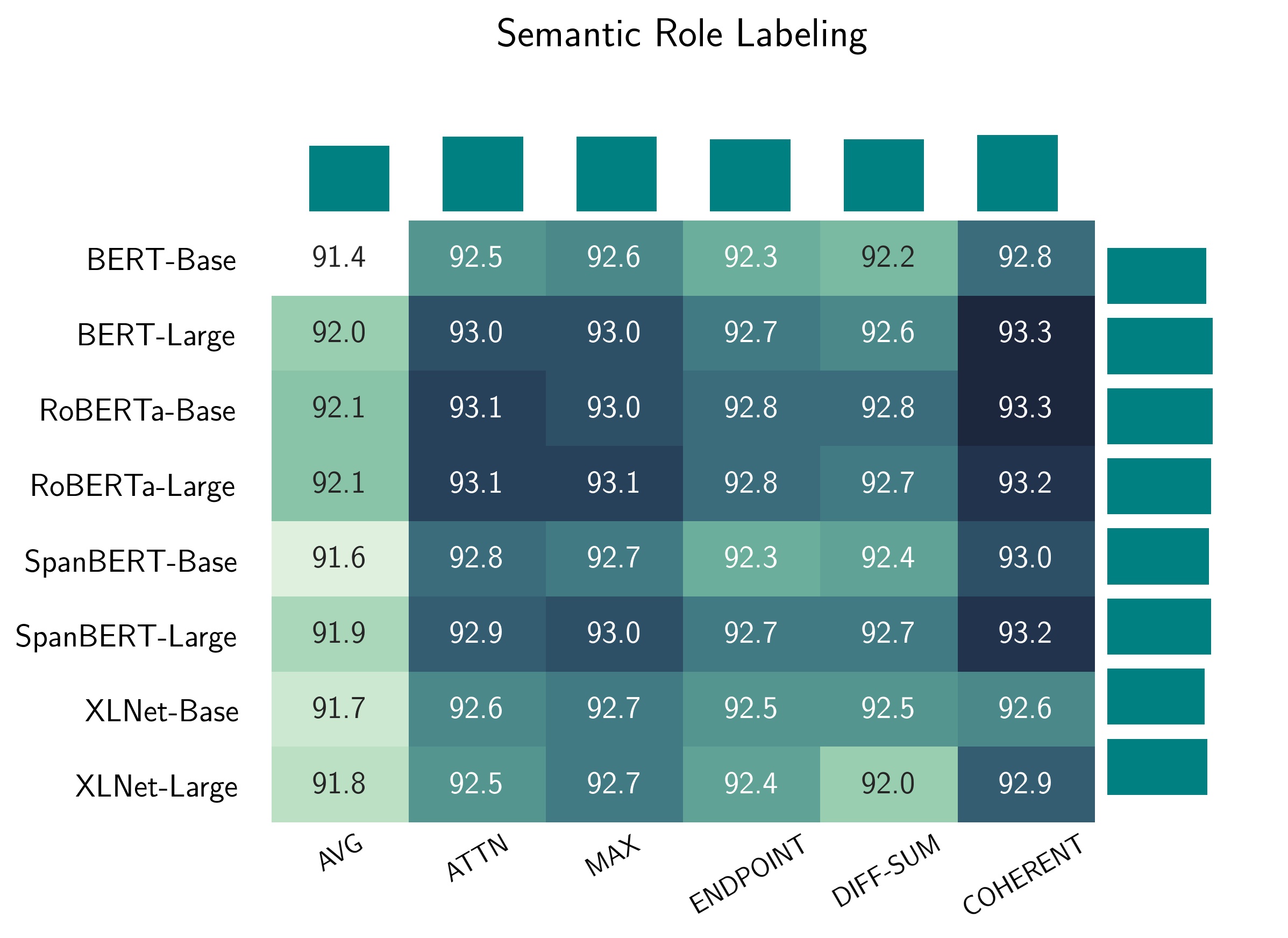}
\end{subfigure}

\medskip
\begin{subfigure}{0.42\textwidth}
 \includegraphics[height=0.18\linewidth, height=2.6in]{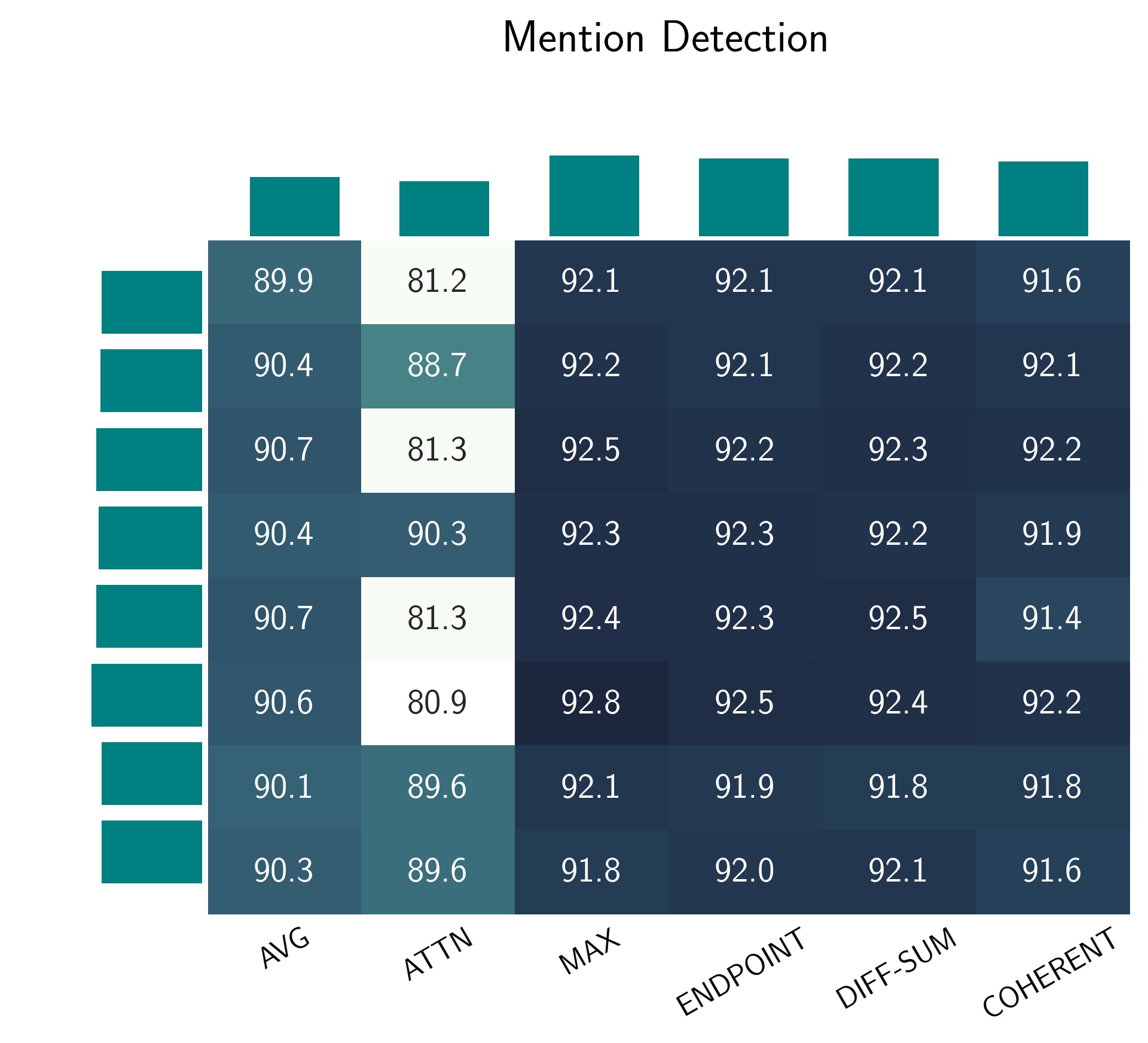}
\end{subfigure}\hspace*{\fill}
\begin{subfigure}{0.56\textwidth}
\includegraphics[height=0.18\linewidth,  height=2.6in]{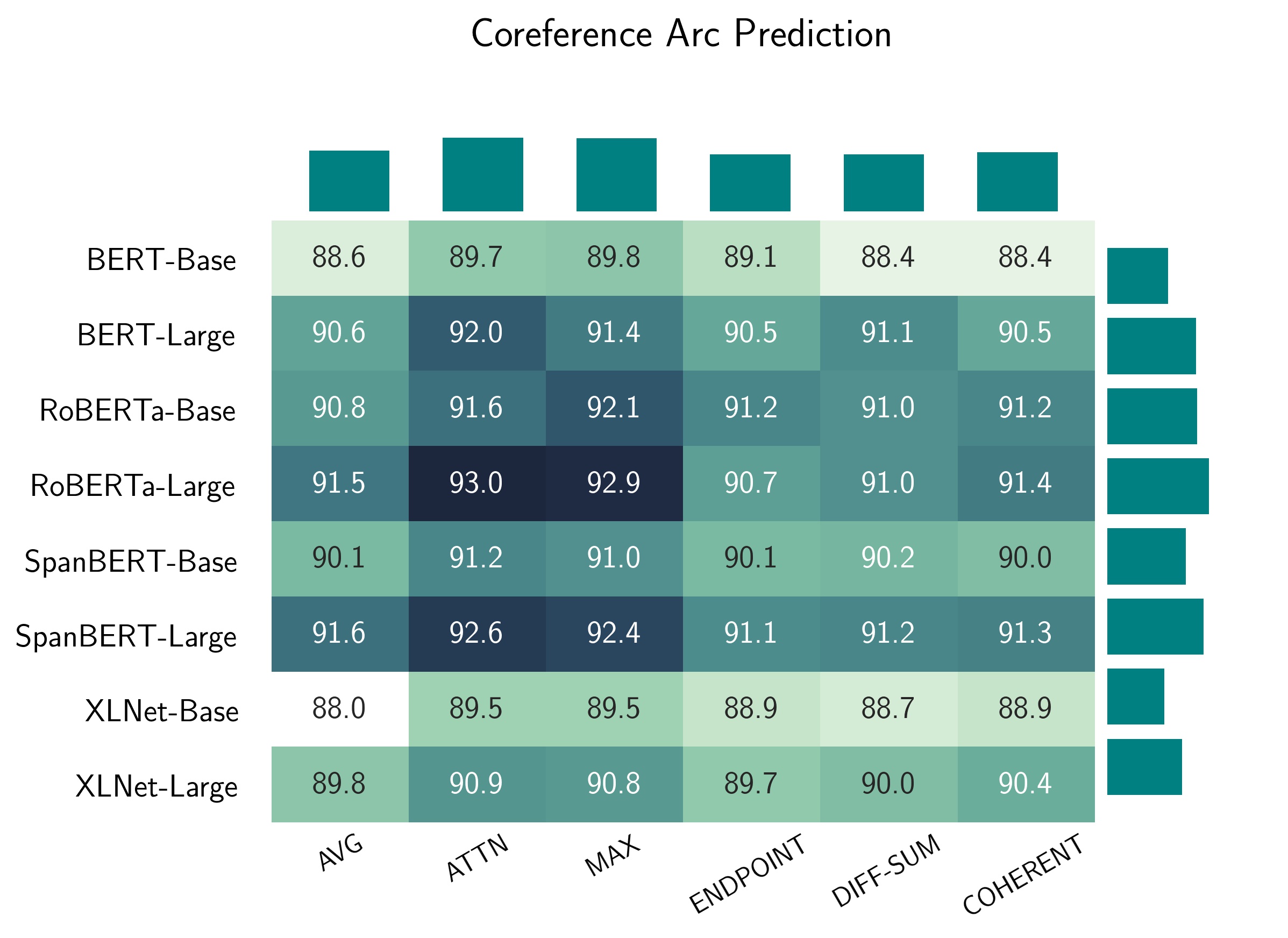}
\end{subfigure}

\caption{Results for the no fine-tuning scenario for the 6 different tasks presented as separate heatmap figures. Each heatmap represents the 48 combinations resulting from 6 span representations and 8 pretrained models. The bars at the side of the heatmap represent the max value in the row/column which is right below the bar. \label{fig:test_perf}} 
\end{figure*}

For the full-scale experiments with fixed pre-trained encoders, we only learn the layer mixing weights. All models are trained using the Adam optimizer \cite{Kingma2015AdamAM} with an initial learning rate of $5 \times 10^{-4}$
and a batch size of 64.\footnote{We found non-trivial gains with this choice of higher learning rate compared to $1\times 10^{-4}$ used by~\citet{tenney2019}.} The model is evaluated on the validation set every 1000 steps and the learning rate is reduced by a factor of 2 if no improvement is seen in the previous 5 validation evaluations. Training stops if no improvement is seen for 20 validation evaluations.

For the fine-tuning experiments, we focus on only a subset of the full-scale configurations for computational reasons. In particular, we only experiment with the ``base" versions of BERT, RoBERTa, and SpanBERT.\footnote{We skipped XLNet due to relatively poor performance across tasks for the non fine-tuned setting.}
All the models are trained using Adam with an initial learning rate of $3 \times 10^{-5}$ and a batch size of 64.
Finally, the token embedding can be a layer-weighted combination or just the last layer.\footnote{Typically the last layer embeddings perform slightly better but a few of those training runs failed and we present the layer-weighted results for those.}

\begin{figure*}[t!] %
\begin{subfigure}{0.42\textwidth}
 \includegraphics[height=2in]{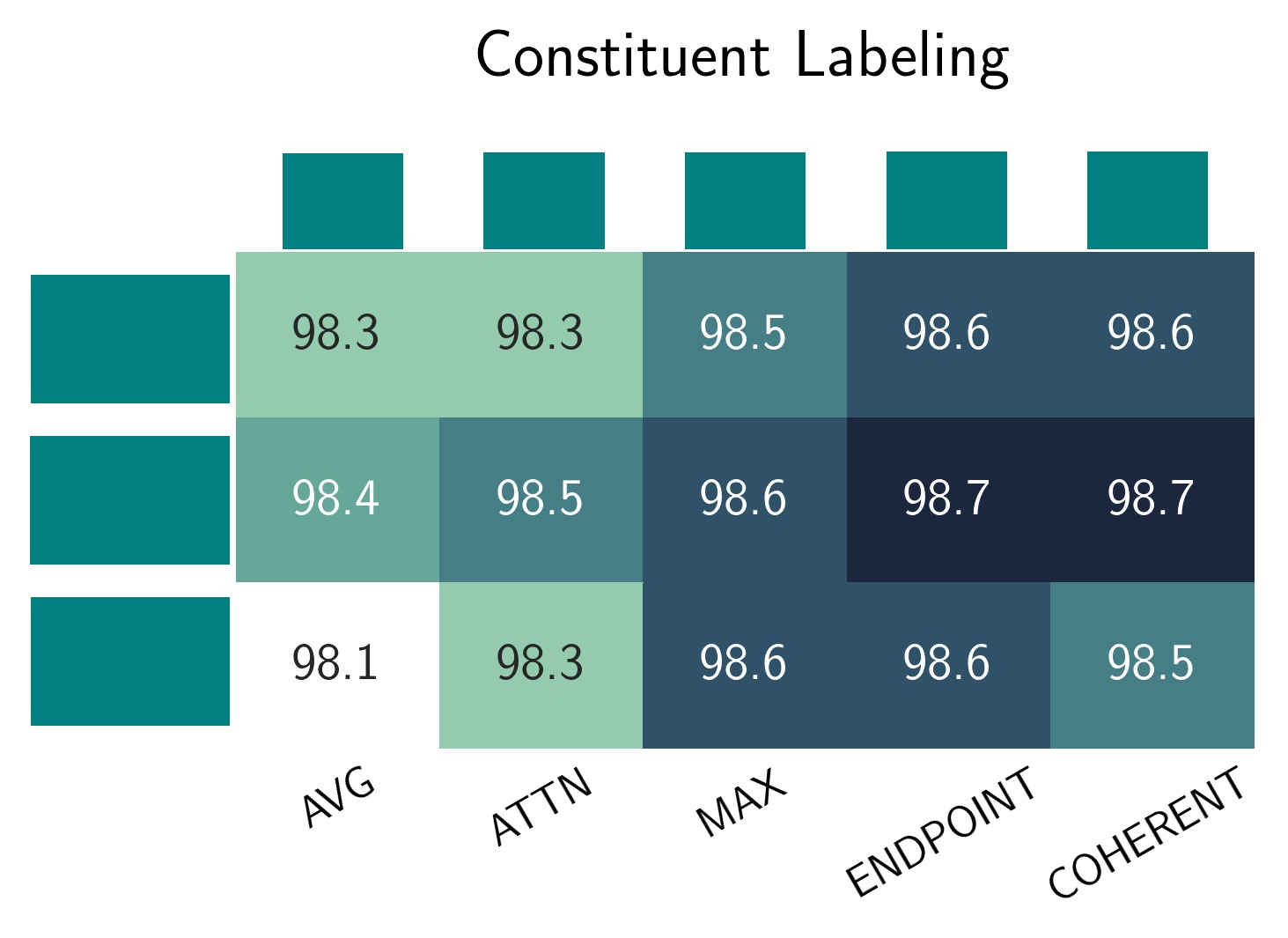}
\end{subfigure}\hspace*{\fill}
\begin{subfigure}{0.56\textwidth}
\includegraphics[height=2in]{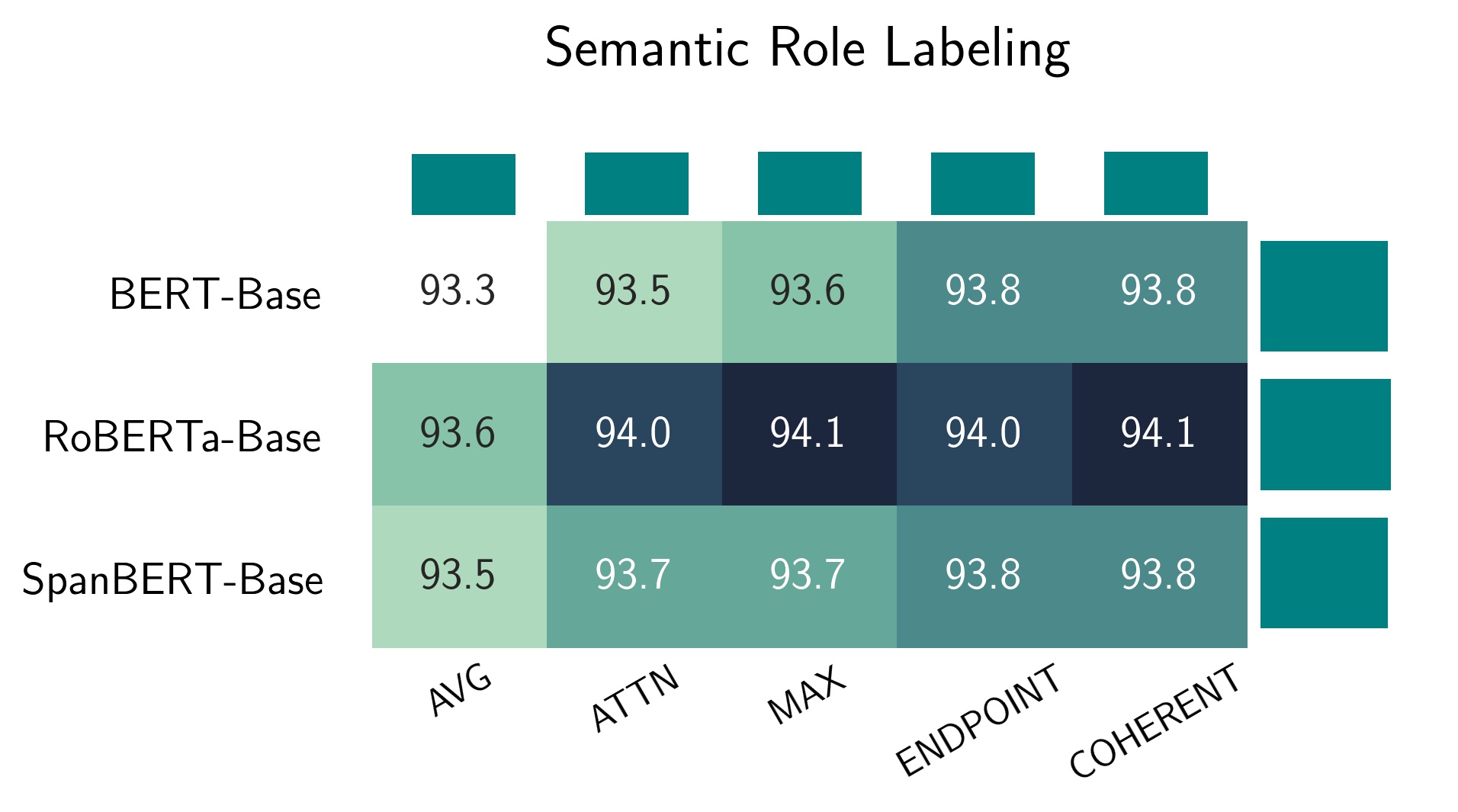}
\end{subfigure}

\smallskip
\begin{subfigure}{0.42\textwidth}
\includegraphics[height=2in]{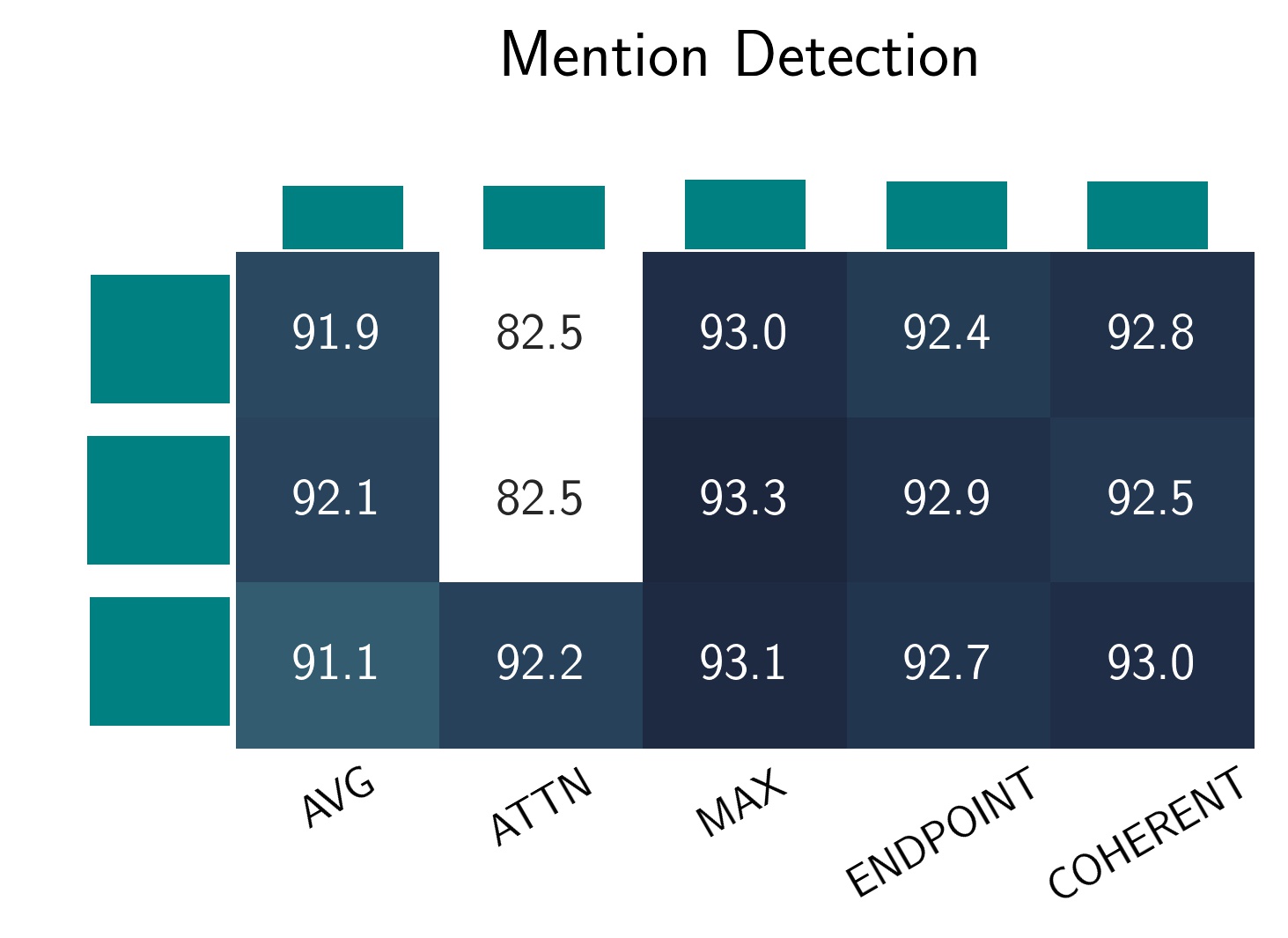}
\end{subfigure}\hspace*{\fill}
\begin{subfigure}{0.56\textwidth}
\includegraphics[height=2in]{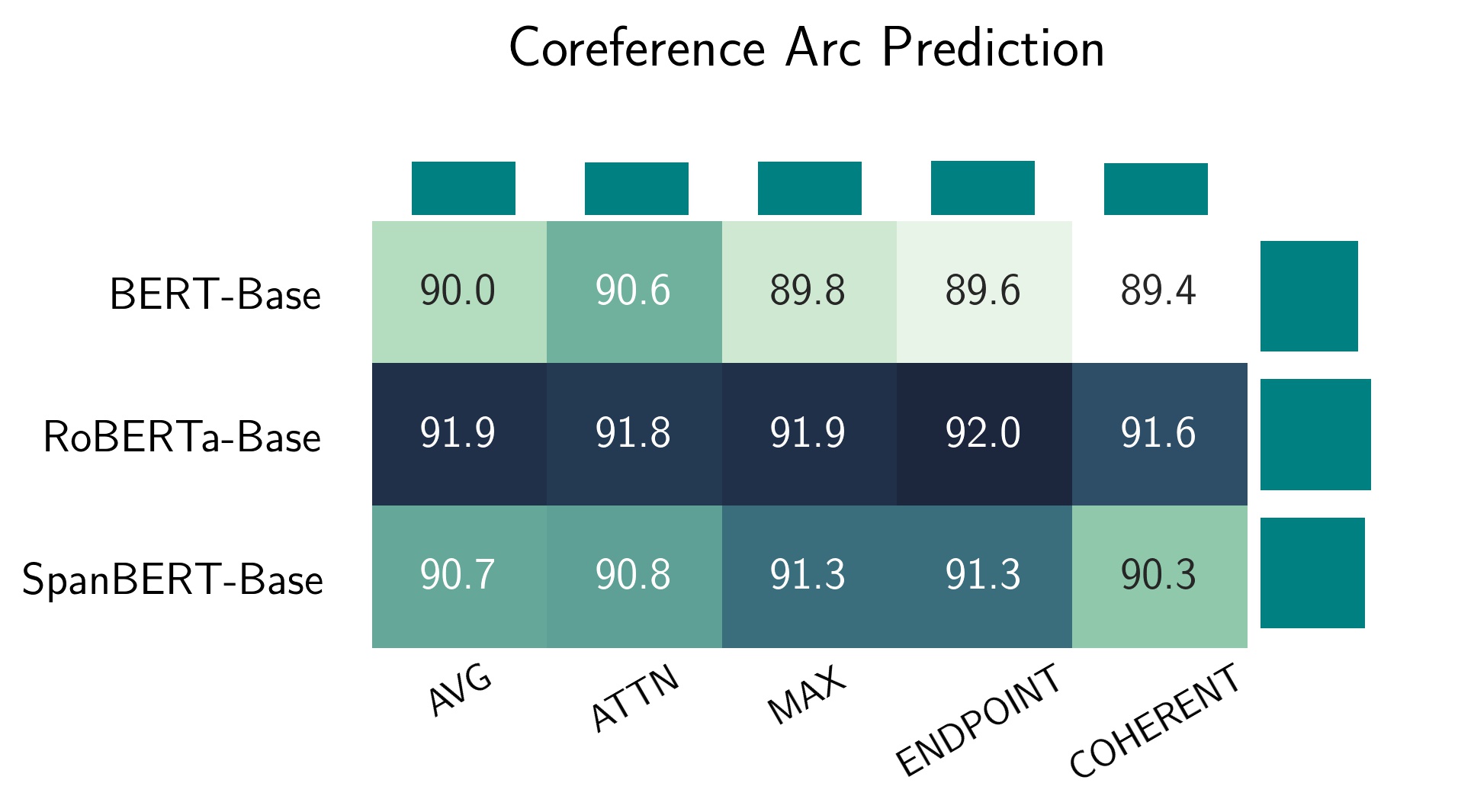}
\end{subfigure}

\caption{Results with fine-tuned encoders for four tasks, namely constituent labeling, SRL, mention detection, and coreference arc prediction, presented as separate heatmaps. \label{fig:ft_test_perf}}
\vspace{-0.1in}
\end{figure*}

\subsection{Data}
Table~\ref{tab:data_stats} shows the dataset statistics of the six tasks evaluated in this study.
For SRL, NEL, coreference arc prediction and constituent labeling, we use the annotations in the OntoNotes 5.0 corpus \cite{ontonotes} and cast the original annotations into the edge probing format, following the same procedure as \citet{tenney2019} for pre-processing.

For the newly proposed constituent detection and mention detection tasks, we create our own datasets using the existing annotations and random negative sampling.
For constituent detection, we use the constituent labeling annotations to get actual constituents, and for each constituent we sample a random negative span of the same length. We ensure that all negative spans are different and that we don't sample an actual constituent.
We follow a similar procedure to get mention detection annotations from coreference arc prediction annotations. To make the mention detection task harder and more realistic, we sample 5 times more negatives than actual mentions.

\section{Results}

\subsection{No Fine Tuning}
\label{sec:no-ft}
The results across tasks and models are shown in Figure \ref{fig:test_perf}. Overall we find max pooling to be the most robust and effective choice across tasks. %
Boundary-based span representations (i.e., \segendpoints, \diffsum, \coherent) are superior to entire-span methods (i.e., \attn, \segmax, \segavg) on tasks which are more shallow/syntactic (e.g., constituent labeling and constituent detection), though max pooling is competitive with the boundary-based methods.

On the other hand, entire-span representations are good at semantic tasks like coreference arc prediction.
As SRL has both semantic and syntactic characteristics, \coherent, \segmax, and \attn show similar performance with the other methods fairly close behind. We do not find large differences between span representation methods for NEL, which mainly contains short spans. %

Model-wise, large models are usually better than base models though there exist exceptions (e.g., constituent labeling). RoBERTa shows strong performance across tasks. We also find that SpanBERT excels for tasks where boundary-based methods are superior, which may be because it is explicitly trained with an objective of predicting tokens inside a span given the boundary tokens. %

Results for each task are summarized below.
\\[0.2cm]
\textbf{Constituent detection/labeling.} Boundary-based representations are better than entire-span ones, though \segmax is close behind.
Surprisingly, in these two tasks, large models are not as good as their base counterparts (\citet{goldberg2019assessing} found similar exceptions for syntactic tasks).\\[0.2cm]
\textbf{Semantic role labeling}. \coherent is the best method on this task with \segmax and \attn  being very close behind.  \\[0.2cm]
\textbf{Mention detection and Coreference arc prediction} \attn and \segmax  perform the best for coreference arc prediction since they benefit from access to the entire span and thus
to the semantic head of the span%
~\cite{lee-etal-2017-end}.
For mention detection the trends are reversed, except for \segmax, with the boundary-based methods doing quite well. This is not surprising since the mention detection task is somewhat close to constituent tasks.
Surprisingly, \attn shows high variance across models and performs worse than even \segavg.
Our initialization of null vector for attention weight means that not learning the attention weight is ironically better than learning it.\footnote{Stopping the gradient for the attention parameter indeed performed similarly to average pooling.}
Preliminary investigation of the learned attention weights didn't give any clues.

Mention detection and coreference arc prediction together complete the pipeline for coreference resolution. The preference for different forms of span representations between the two (except for \segmax) suggests that different span representations can be considered for different stages of the coreference resolution task. Interestingly, one of the best performing end-to-end coreference models~\cite{lee-etal-2017-end} uses a concatenation of a boundary-based span representation, \segendpoints, and \attn.

Some of our observations may be confounded with training set sizes, which vary from coreference arc prediction on the small end (208K) up to constituent tasks on the largest end (1.9M), with SRL (599K) in the middle of the range.

\begin{figure*}[ht]
\centering
\begin{minipage}{.36\textwidth}
  \centering
  \includegraphics[width=\linewidth, height=1.6in]{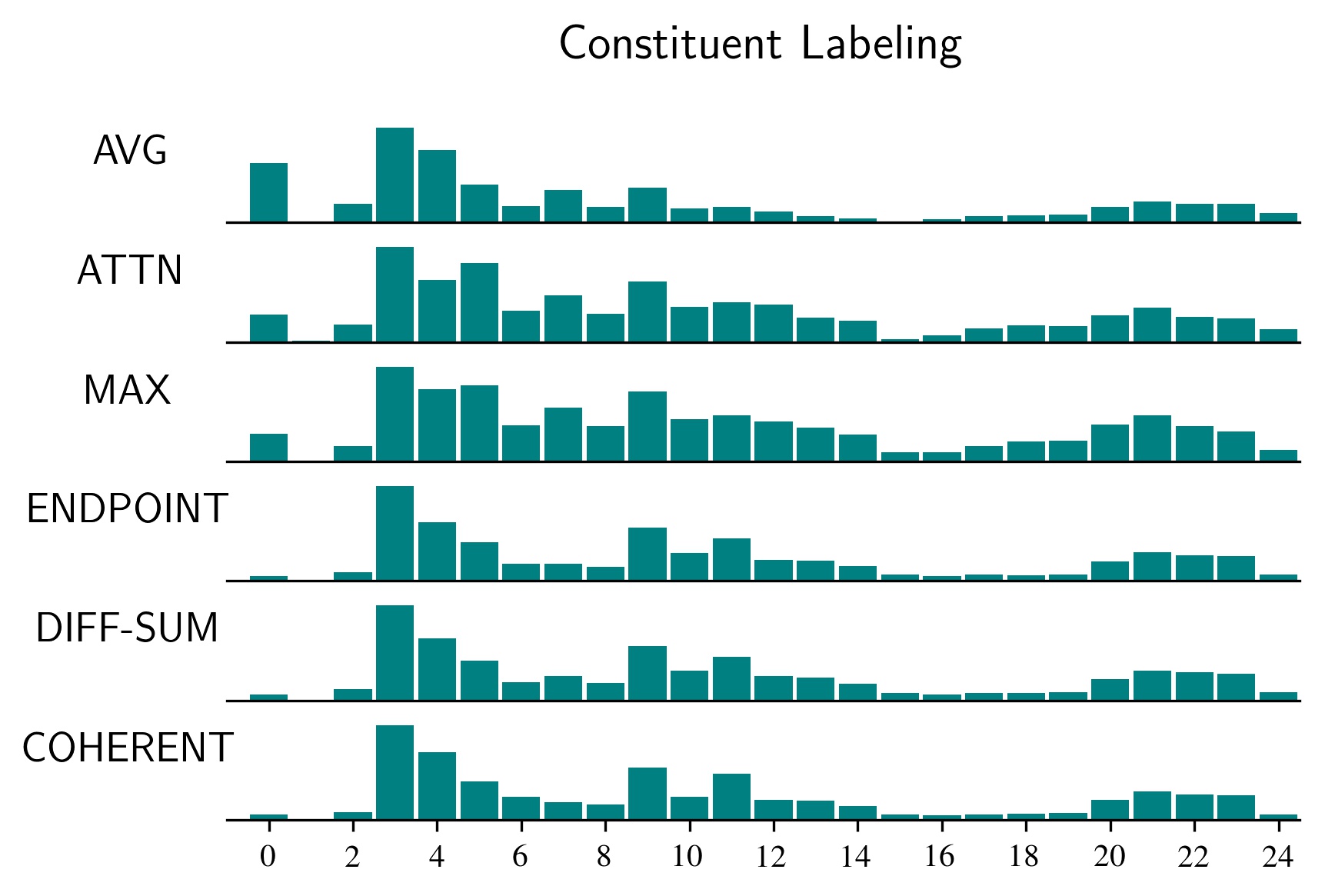}
\end{minipage}%
\begin{minipage}{.32\textwidth}
  \centering
  \includegraphics[width=\linewidth, height=1.6in]{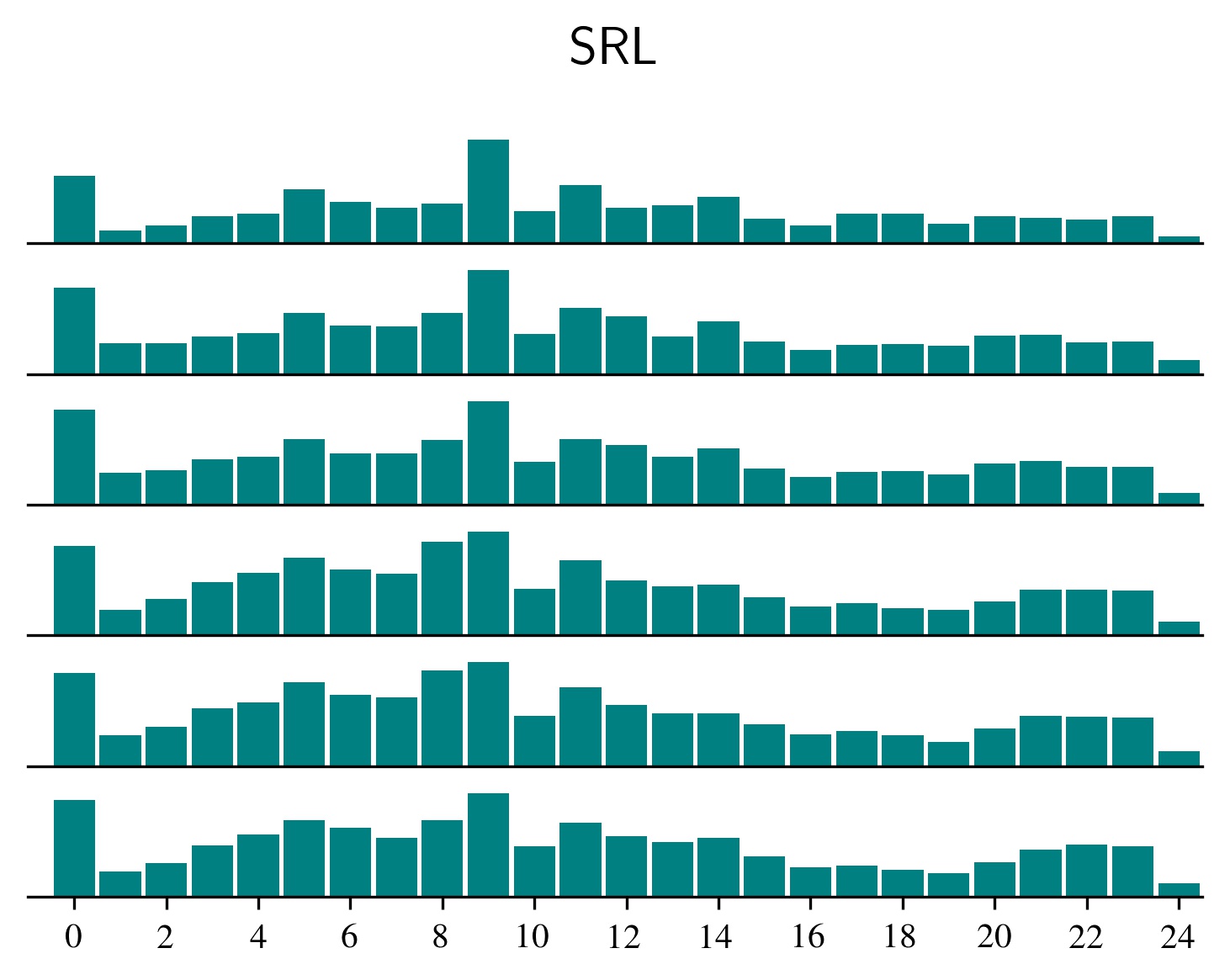}
\end{minipage}%
\begin{minipage}{.32\textwidth}
  \centering
  \includegraphics[width=\linewidth, height=1.6in]{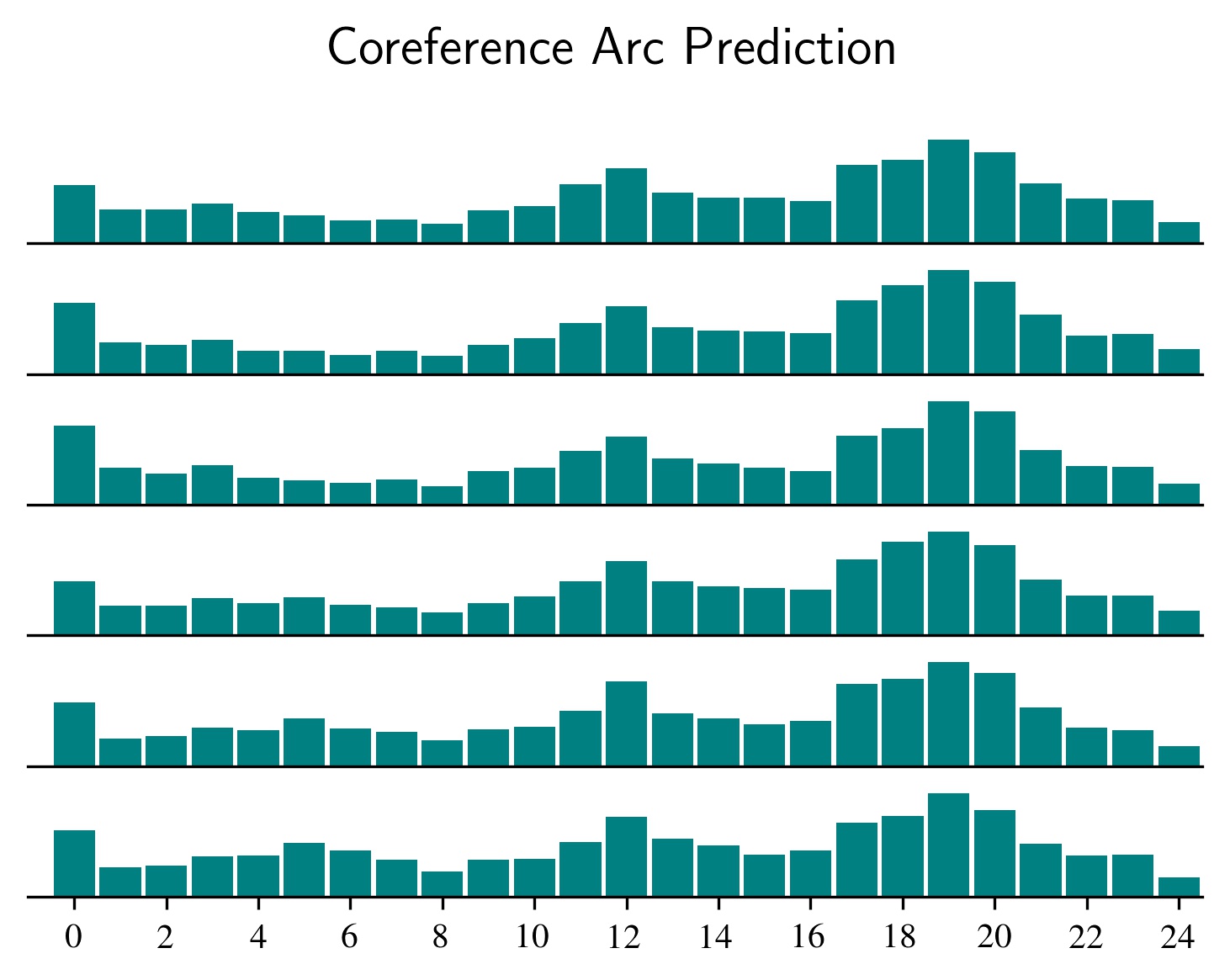}
\end{minipage}
\caption{Visualization of layerwise weights learned for all the span representations for constituent labeling, SRL, and coreference arc prediction for the RoBERTa-large model.).}%
\label{fig:layerweights_1}
\end{figure*}

\begin{figure*}[ht!]
\centering
\begin{minipage}{.36\textwidth}
  \centering
  \includegraphics[width=\linewidth, height=1.5in]{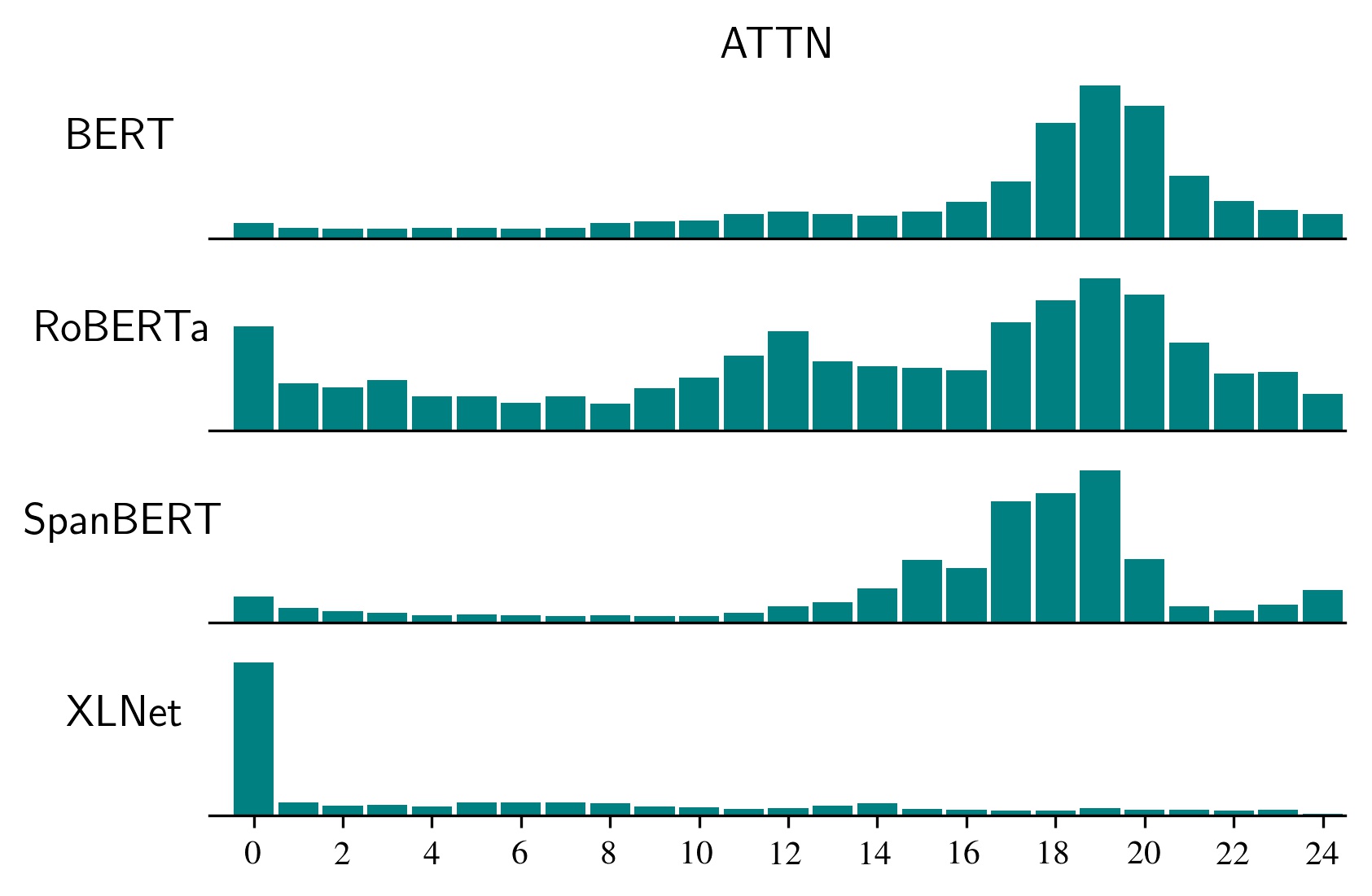}
\end{minipage}%
\begin{minipage}{.32\textwidth}
  \centering
  \includegraphics[width=\linewidth, height=1.5in]{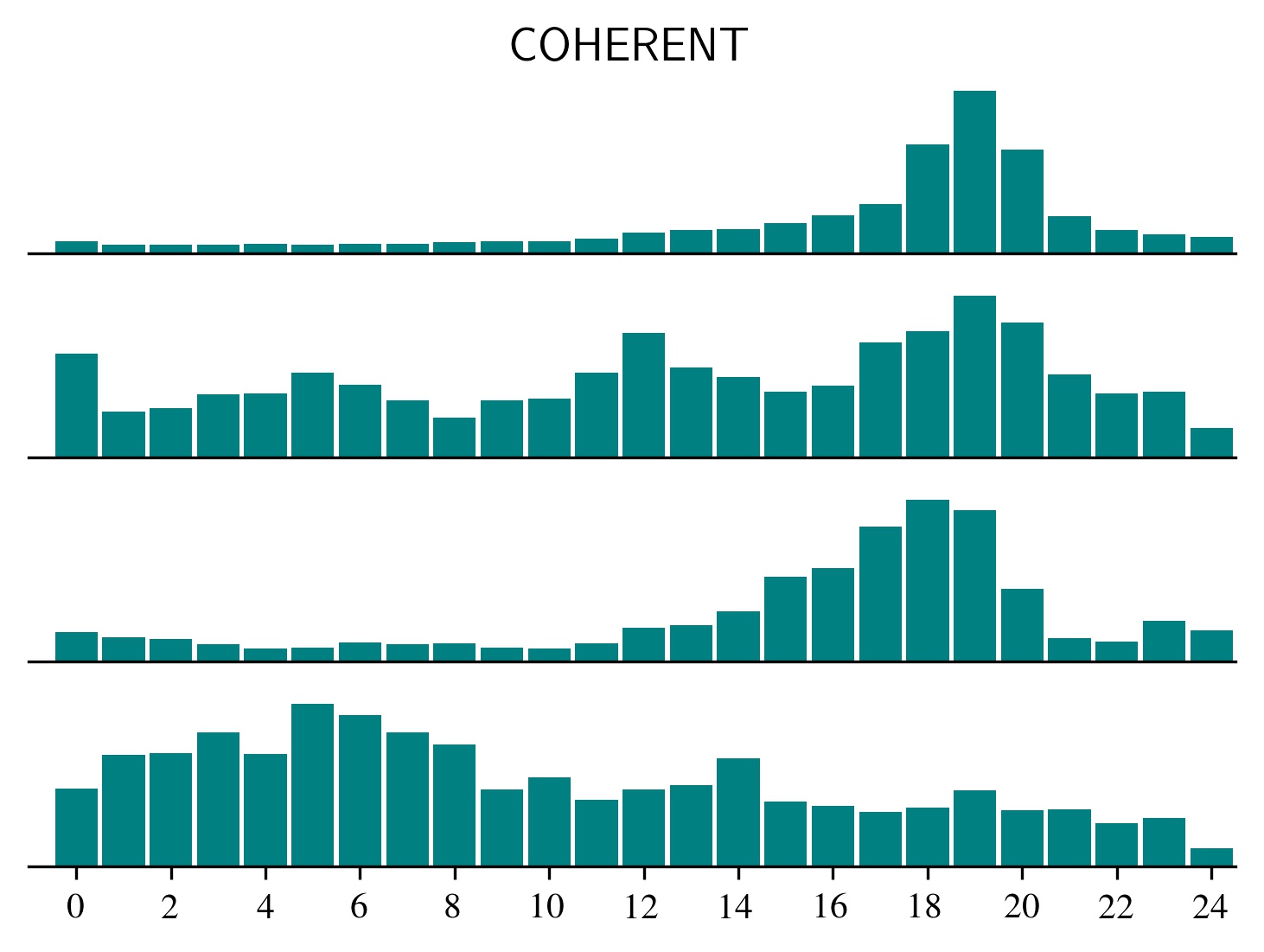}
\end{minipage}%
\begin{minipage}{.32\textwidth}
  \centering
  \includegraphics[width=\linewidth, height=1.5in]{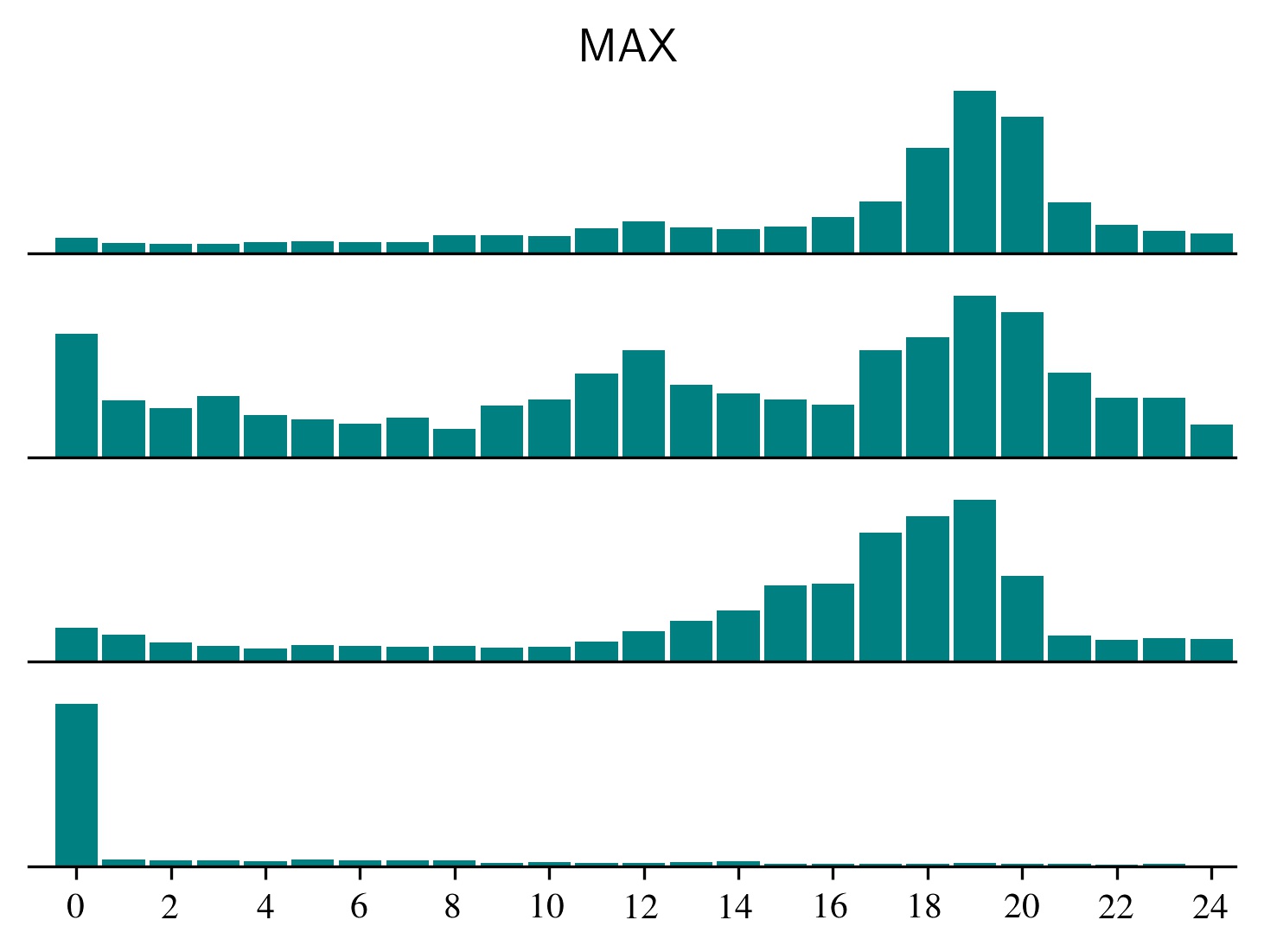}
\end{minipage}
\caption{Visualization of layerwise weights learned for the coreference arc prediction task for the combination of span representations ATTN, COHERENT, and MAX with \{BERT, RoBERTa, SpanBERT, XLNet\}-large models. While BERT, SpanBERT, and XLNet have peaky weights, RoBERTa's weights are more spread out. Oddly enough XLNet chooses to place the most weight on the embedding layer for ATTN and MAX (the two best span representations for XLNet-large).}
\label{fig:layerweights_2}
\end{figure*}

\newcommand{\specialcell}[2][t]{%
  \begin{tabular}[#1]{@{}l@{}}#2\end{tabular}}
\newcommand{\predicat}[1]{\underline{#1}}
\newcommand{\argument}[1]{\textbf{#1}}

\begin{table*}[t]
\setlength{\tabcolsep}{4pt}
\centering
\small
\begin{tabular}{p{1.8cm}cp{4.3cm}p{8.1cm}}
\toprule
Argument & $\Delta$R & Notes & Examples \\
\midrule
\specialcell{C-ARG1\\(continuous\\argument)}
& 10.85 & often starts with \emph{to}, similar to \texttt{xcomp} in universal dependencies & ``were \predicat{granted} the right earlier this year \argument{to ship sugar}'', ``\predicat{brought} more money into a city \argument{than it took out}'', ``Food prices are \predicat{expected} \argument{to be unchanged}'' \\
\midrule[0.1pt]
ARGM-DIS (discourse) & 3.24 & mostly single-word discourse modifiers like \emph{and} or \emph{but} & ``\argument{In addition} , the government is \predicat{figuring}'', ``\argument{But} \predicat{eluding}'', ``\argument{And} the USIA \predicat{said}'', ``\argument{of course} , there \predicat{'s} that word'' \\
\midrule[0.15pt]
ARGM-MNR (manner) & 2.03 & mostly adverbs and prepositional phrases &
``...get \predicat{married} \argument{in a tuxedo}'', ``\argument{relatively} \predicat{respected}'', ``\predicat{Moving} \argument{rapidly} through school'',
``...do n't \predicat{leave} home \argument{without the American Express card}''
\\
\midrule[0.1pt]
\specialcell{ARG3\\ (starting point,\\ benefactive,\\ attribute)} & -4.28 & multiple functions depending on predicate, and thus a variety of boundary words  &
``And what I had \predicat{mentioned} \argument{about my mother bugging me} was...'', %
``You should feel comfortable \predicat{staying} \argument{there}'', %
``...he believes that it can \predicat{bring} the market \argument{back up} after a plunge'',
``Biologists \predicat{mixed} a mold element in the cells of plants with pearl powder \argument{to produce a granulated drug}''
\\
\midrule[0.1pt]
ARGM-DIR (direction) & -5.20 & typically a word or short phrase, mostly adverbs, prepositional phrases, particles, and adjectives  &
``newspapers \predicat{turning} \argument{to color} on their pages'',
``bond prices rapidly \predicat{turned} \argument{south}'',
``major brokerage firms \predicat{rushed} \argument{out} ads'',
``\predicat{takes} the dispute \argument{to the Supreme Court}'',
``we have to \predicat{get} \argument{out of bed}'',
``\predicat{toss} the chalk \argument{back and forth}''
\\
\midrule[0.1pt]
ARGM-EXT (extent) & -8.14 & %
often a short phrase like \emph{more}, \emph{very much}, \emph{a lot}, etc.; limited semantics, range of surface forms %
& ``you \predicat{'re} critical to yourself \argument{too much}'', ``of \predicat{freezing} , \argument{at least partially}'', ``\predicat{increase} \argument{of 32 \%}'', ``life has \predicat{changed} \argument{a lot}'', ``\predicat{Thank} you \argument{very much}''
\\
\bottomrule
\end{tabular}
\caption{
Analysis of argument labels for semantic role labeling. $\Delta$R~=
argument recall\% with boundary-based span representation methods minus recall\% with entire-span methods.
In the examples, predicates are underlined and arguments of the given type are shown in boldface.
\label{tab:analysis-srl}
}
\vspace{-0.1in}
\end{table*}

\subsection{Fine Tuning}
State-of-the-art models almost always fine-tune the pretrained encoders.
However, the training is quite computationally expensive; hence, we perform the span representation comparison for a small set of the total configurations whose results are shown in Figure~\ref{fig:ft_test_perf}. In general, fine tuning improves the result for all span representation methods across tasks with the performance of different span representation methods more tightly clustered with fine-tuning in comparison to no fine-tuning. This is best illustrated by the constituent labeling task where without fine-tuning average pooling trails by 10+ F-score with respect to the best span representation but only less than 0.5 F-score with fine-tuning.

\section{Analysis}
\label{sect:analysis}

In this section we analyze the impact of span representation method with fixed pretrained encoders.
\subsection{Layerwise Weight Analysis}
\label{sec:analysis:layerweights}
Figures~\ref{fig:layerweights_1} and ~\ref{fig:layerweights_2} visualize learned layer weights for different task, span representation, and pretrained encoder combinations. %
Within a model and task, we generally found that the layer weights were fairly consistent across span representation methods. Overall, we find  similar trends to prior work in analyzing layer weights for downstream NLP tasks, namely that constituency parsing has higher weight for lower layers and coreference has most weight on higher layers, with SRL in between~\citep{liu-etal-2019-linguistic,tenney-etal-2019-bert}. For ablation analysis of layer weights, whether to learn them or not, see Appendix~\ref{sec:app_mix_wt}.

\subsection{Label-Specific Analysis of Span Groups}
\label{sect:label_analysis}
We seek to determine whether boundary-based span representation methods (\coherent, \diffsum, and \segendpoints) differ systematically from methods that consider the entire span (\attn, \segmax, and \segavg). %
We pooled predictions from the three methods in the group for \robertalarge and calculated the recall for particular labels, for two tasks: SRL, and constituent labeling (analysis for NEL in Appendix~\ref{sect:app_label_nel}).
We found the labels with the largest differences in recall between the two groups, and discuss our findings below.%

\paragraph{SRL.}
Table~\ref{tab:analysis-srl} shows the argument labels with the largest differences in recall ($\Delta$R) between the two groups, limiting our analysis to the 20 most frequent argument labels. Labels with positive $\Delta$R~are handled better by boundary-based methods. These tend to be arguments that can be identified based on particular words, often function words, at the boundaries. For example, ARGM-DIS is found mostly with single-word modifiers in this dataset (like \emph{and} and \emph{but}). %
Arguments that are handled better by entire-span methods are more diverse in terms of their boundary words. %
ARGM-EXT is used for arguments with relatively limited semantics (as shown in the examples) but a variety of surface realizations.

\begin{table}[t]
\setlength{\tabcolsep}{4pt}
\centering
\small
\begin{tabular}{p{1.1cm}cp{5.1cm}}
\toprule
Label & $\Delta$R & Examples \\
\midrule
SBAR & 6.1 & ``that are missing'', ``who owned the land'' \\
PRN & 4.6 & ``, she says ,'',
``, it turns out ,'', ``( file photo )'',
``( hey , it 's possible )''\\
ADJP & 3.3 & ``liable'', ``available to anyone'', ``more generous'', ``satisfied with where they work'', ``at least somewhat interesting'' \\
PP & 2.7 & ``in 1966'', ``within a community'' \\
\midrule
SBARQ& -6.1 & ``What can we do ?'', ``So what should be done .'', ``and what is money for'', ``how shall I say'' \\
SQ & -6.5 & ``Did you see ?'', ``will I do now'', ``do you make of'', ``You still building'' \\
FRAG & -6.5	& ``Or something .'', ``well below 1988 activity'', ``As for Mr. Papandreou ?'' \\
SINV & -15.7 & ``should the Air Force order the craft'', ``say Mr. Dinkins 's managers'', ``notes Huang frankly'', ``invest they will''  \\
\bottomrule
\end{tabular}
\caption{
Analysis of labels for constituent labeling. $\Delta$R~=
label recall\% with \segendpoints minus recall\% with \segmax. We restrict this analysis to labels that appear at least 100 times in our development set.
\label{tab:analysis-constlabel}
}
\vspace{-0.1in}
\end{table}

\paragraph{Constituent Labeling.}
Table~\ref{tab:analysis-constlabel} shows a similar analysis for constituent labeling, though in this case we compare only a single method from each family: \segendpoints and \segmax. We do this because \segmax is comparable in performance to the boundary methods while \attn and \segavg are significantly worse. We choose \segendpoints as our single representative of the boundary methods in order to compare only two methods, though we found the same trends for others in its group.

\segendpoints has higher recall on several labels, shown in the top part of the table. There is a 6\% difference for SBAR, which is a clause introduced by a (possibly empty) subordinating conjunction. About 25\% of SBAR constituents begin with \emph{that}, and many others start with some other very common subordinating conjunction, making SBAR easier to find for methods that focus on boundary words. Parentheticals (PRN) frequently begin and end with commas or parentheses. ADJPs typically begin or end with an adjective and PPs nearly always begin with prepositions.

The lower part of Table~\ref{tab:analysis-constlabel} shows labels where \segmax has higher recall than \segendpoints. The largest difference is in SINV, which is an ``inverted'' declarative sentence, that is, a sentence in which the subject follows the conjugated verb. These often look like VPs based on boundary words but are more diverse syntactically; a few short examples are shown in the table. The other labels also show syntactic diversity. FRAG (fragment) has many realizations that vary widely in terms of their syntax. while SBARQ and SQ often start with \emph{wh}-words and end in question marks, they show significant variation.

\section{Related Work}

Many of the span representations that we consider here were proposed previously for specific tasks, such as the attention-weighted pooling of \citet{lee-etal-2017-end} for coreference resolution; the endpoint-based  representation of \citet{lee-etal-2016-learning} and the ``coherent" endpoint-based representation of \citet{seo-etal-2019-real} for question answering; and combinations of differences and sums of endpoint representations for parsing and semantic role labeling \citep{stern-etal-2017-minimal,ouchi-etal-2018-span}. These are described in more detail in Section~\ref{sec:pooling_models}.  

Other recent work has considered pooling approaches such as the difference between endpoint representations \citep{wang-chang-2016-graph,cross-huang-2016-span} or a concatenation of endpoint and attention-based representations \citep{li-etal-2016-discourse}. Other approaches concatenate additional specialized feature vectors, such as the span length or position information \citet{lee-etal-2017-end, he-etal-2018-jointly, kuribayashi-etal-2019-empirical}.
Some work has also considered explicitly composing span representations via syntactic parse trees, such as recursive neural networks \citep{li-etal-2014-recursive}, and some unsupervised parsing models produce span representations as a byproduct of training \citep{drozdov-etal-2019-unsupervised-latent}. 

At the same time, there has been significant effort devoted to the related problem of learning representations for sentences or even longer texts \interalia{kalchbrenner2014convolutional,iyyer2015deep,kiros2015skip,wieting-16-full,conneau-etal-2017-supervised,Shen2018Baseline}.  Much of this work focuses on pooling over word representations, often finding that simple pooling operations like averaging perform surprisingly well \citep{wieting-16-full,Shen2018Baseline}.
 \citet{Shen2018Baseline} did a similar empirical study to ours in spirit, comparing a variety of pooling models for sentence representations across tasks.

In this work we are mainly focusing on the models for computing span representations given pretrained token embeddings, but we also include a variety of pretrained contextual embeddings.  One in particular, SpanBERT \citep{joshi2019spanbert},   was designed to enable improved span representations. While recent work has compared across pretrained contextual embeddings for representing spans \citep{tenney2019}, to our knowledge there has been no systematic comparison of methods for combining these contextual embeddings into span representations across a variety of tasks.

\section{Conclusion}
We systematically compare multiple span representation methods, %
 combined with various base embedding models, on various tasks.  Our analysis includes two new tasks that we propose to tease apart different aspects of span representations. For the fixed pre-trained encoder, we find that, although max pooling is a fairly reliable representation across tasks, the optimal span representation varies with respect to the syntactic and semantic nature of different tasks.
Finally, fine-tuning greatly reduces the impact of span representation choice on performance.

\bibliography{acl2020}
\bibliographystyle{acl_natbib}
\appendix
\section{Appendix}

\subsection{Mixing Weights for Layers}
\label{sec:app_mix_wt}
\begin{table}[htbp]
\setlength{\tabcolsep}{4pt}
\small{
\begin{tabular}{lllllll}
\toprule
Model & AVG & ATT & MAX & EP & DS & COH\\
\midrule
BERT-large     & 90.6 & 92   & 91.4 & 90.5 & 91.1 & 90.5 \\
\nlw & 90.3 & 91.3 & 91.5 & 90.4 & 90.4 & 90.3 \\
RoBERTa-large  & 91.5 & 93   & 92.9 & 90.7 & 91.0 & 91.4 \\
\nlw & 91.0 & 92.7 & 92.6 & 90.9 & 90.7 & 90.6 \\
SpanBERT-large & 91.6 & 92.6 & 92.4 & 91.1 & 91.2 & 91.3 \\
\nlw & 90.4 & 91.4 & 91.5 & 90.6 & 90.3 & 90.2 \\
XLNet-large    & 89.8 & 90.9 & 90.8 & 89.7 & 90.0 & 90.4 \\
\nlw & 89.4 & 90.7 & 90.8 & 89.5 & 90.0 & 90.7 \\
\bottomrule
\end{tabular}
}
\caption{Analysis of importance of learning mixing weights for combination of different models and span representations for the coreference arc prediction task.
}
\end{table}

In the table above we analyze the effect of learning the layerwise mixing weights vs simple averaging over layers in context of the coreference arc prediction task. \attn-based models suffer the biggest drop with a drop of 0.6\% absolute on average. Among pretrained contextual embedding models, SpanBERT-large is hurt the most with a drop of 1\% absolute on average.
Surprisingly, XLNet drops by only 0.1\% on average even though its attention plots looked quite peaky for some of the span representations.

\subsection{Label-Specific Analysis of Span Groups for NEL}
\label{sect:app_label_nel}
\begin{table}[htbp]
\setlength{\tabcolsep}{4pt}
\centering
\small
\begin{tabular}{p{1.4cm}cp{4.9cm}}
\toprule
Label & $\Delta$R
& Examples \\
\midrule
ORDINAL & 1.2
& ``first'', ``second'', ``First'', ``6th'', ``ninth'' \\
CARDINAL & 0.3
& ``two'', ``10'', %
``Dozens'', ``at least 37'' \\ %
\midrule
TIME & -2.3 & ``seven o'clock'', ``two hours'', ``about ten'', ``eight fifty in the morning'' \\
LAW & -2.9 & ``Paragraph 14 of Article 19'', ``the Geneva Convention'', ``Dru 's Law'' \\
LOC & -3.2 &
``the Sierra Nevada Mountains'', ``Asia'', ``Mai Po Marshes'' \\
WORK\_ OF\_ART & -6.0
& ``The End of the Day'', ``Carry On Trading'', ``News Night Tonight'' \\
\bottomrule
\end{tabular}
\caption{
Analysis of labels for NEL. $\Delta$R~=
label recall\% with boundary-based span representation methods minus recall\% with entire-span methods.
\label{tab:analysis-entity}
}
\end{table}

\paragraph{NEL.} Table~\ref{tab:analysis-entity} shows a similar analysis for entity labeling as done in Section~\ref{sect:label_analysis}. The labels with higher recall under the boundary-based methods are limited to ORDINAL and CARDINAL numbers, which tend to be very short and highly regular (nearly all ORDINAL entities are one token and approximately half are \emph{first}). The entire-span methods achieve much higher recall for the WORK\_OF\_ART label, and also for LOC, LAW, and TIME. These entities tend to be multi-word phrases with a variety of syntactic forms and without consistent boundary words.

It may be surprising that ORDINAL is better detected by the boundary methods, since nearly all ORDINAL entities are a single token, and the entire-span methods reduce to a simple form for single tokens. However, this may show that the entire-span methods are being trained to abstract over the contents of the span,
thereby losing some of the surface information. The boundary-based methods, by contrast, devote particular parts of the span representation to the boundary position representations, thereby providing a more direct/explicit connection between those boundary words and the downstream classifier.

\end{document}